\documentclass[letterpaper, 10 pt, journal]{ieeeconf}

\IEEEoverridecommandlockouts                              

\usepackage{etoolbox}
\makeatletter
\patchcmd{\@makecaption}
  {\scshape}
  {}
  {}
  {}
\makeatletter
\patchcmd{\@makecaption}
  {\\}
  {.\ }
  {}
  {}
\makeatother

\usepackage{graphics} 
\usepackage{epsfig} 
\usepackage{mathptmx} 
\usepackage{times} 
\usepackage{amsmath} 
\usepackage{amssymb}  
\usepackage{algorithm}
\usepackage{algpseudocode}
\usepackage{cite}
\usepackage{hyperref}
\usepackage{booktabs}
\usepackage{caption}
\usepackage{makecell} 
\usepackage{multirow}
\usepackage{adjustbox}
\usepackage{changepage}
\usepackage{xcolor}
\usepackage{wrapfig}
\usepackage{subcaption}

\usepackage{geometry}
\graphicspath{{Figures/}}

\hypersetup{
  colorlinks   = true,    
  urlcolor     = blue,    
  linkcolor    = blue,    
  citecolor    = blue      
}

\DeclareMathAlphabet{\mathcal}{OMS}{cmsy}{m}{n}

\title{\LARGE \bf Off-dynamics Conditional Diffusion Planners}

\author{Wen Zheng Terence Ng, Jianda Chen, Tianwei Zhang 
\\
{\small Nanyang Technological University, Singapore}\\
{\tt\small\{ngwe0099, jianda001\}@e.ntu.edu.sg, tianwei.zhang@ntu.edu.sg }\\ 
}      

\begin{document}

\maketitle
\thispagestyle{empty}
\pagestyle{empty}

\begin{abstract}

Offline Reinforcement Learning (RL) offers an attractive alternative to interactive data acquisition by leveraging pre-existing datasets. However, its effectiveness hinges on the quantity and quality of the data samples. This work explores the use of more readily available, albeit off-dynamics datasets, to address the challenge of data scarcity in Offline RL.
We propose a novel approach using conditional Diffusion Probabilistic Models (DPMs) to learn the joint distribution of the large-scale off-dynamics dataset and the limited target dataset. To enable the model to capture the underlying dynamics structure, we introduce two contexts for the conditional model: (1) a \textit{continuous dynamics score} allows for partial overlap between trajectories from both datasets, providing the model with richer information; (2) an \textit{inverse-dynamics context} guides the model to generate trajectories that adhere to the target environment's dynamic constraints.
Empirical results demonstrate that our method significantly outperforms several strong baselines. Ablation studies further reveal the critical role of each dynamics context. Additionally, our model demonstrates that by modifying the context, we can interpolate between source and target dynamics, making it more robust to subtle shifts in the environment.

\end{abstract}

\section{Introduction}
Conventional Reinforcement Learning (RL) excels at learning from live interactions within its environment, but real-time data acquisition can be resource-intensive, dangerous or infeasible in critical domains. \emph{Offline RL} emerges as a powerful alternative, leveraging pre-existing datasets without live interactions to train control policies \cite{levine2020offline, lange2012batch}. This approach opens doors to training intelligent agents across a wide range of applications, including life-saving surgical robots \cite{zareleveraging, fan2024learn}, data-driven financial trading \cite{qin2022neorl, lee2023offline}, improved medical diagnosis \cite{tseng2017deep, nie2021learning}, and autonomous vehicles \cite{kendall2019learning, codevilla2018end, sun2018fast}. 
Despite its potential, its effectiveness heavily relies on the quantity and quality of available data, with performance deteriorating sharply when data is scarce \cite{mandlekar2022matters,liu2021dara}. 
However, collecting large-scale expert-level data from offline sources remains challenging despite not requiring live interaction with the environment.

This work explores the use of more accessible source datasets to address the challenge of data scarcity in Offline RL. We draw inspiration from the success of transfer learning in supervised learning \cite{pan2009survey, tan2018survey}, where researchers have increasingly utilized data from additional, easier-to-obtain sources \cite{raffel2020exploring,ruder2019transfer, gopalakrishnan2017deep}. 
In the context of Offline RL, we apply this principle by employing a readily available \emph{off-dynamics} source dataset that aligns with the target task's objective but possesses different transition dynamics compared to the target environment.
This strategy can be well applicable to real-world settings. For instance, self-driving cars can leverage data from diverse cities or vehicle models; medical diagnosis can benefit from similar but non-identical illness data; surgical robots can initially train on artificial organs; financial trading algorithms can utilize data from larger, related markets to inform decisions in smaller ones.

\begin{figure}[t]
    \centering
    \includegraphics[width=0.99\linewidth]{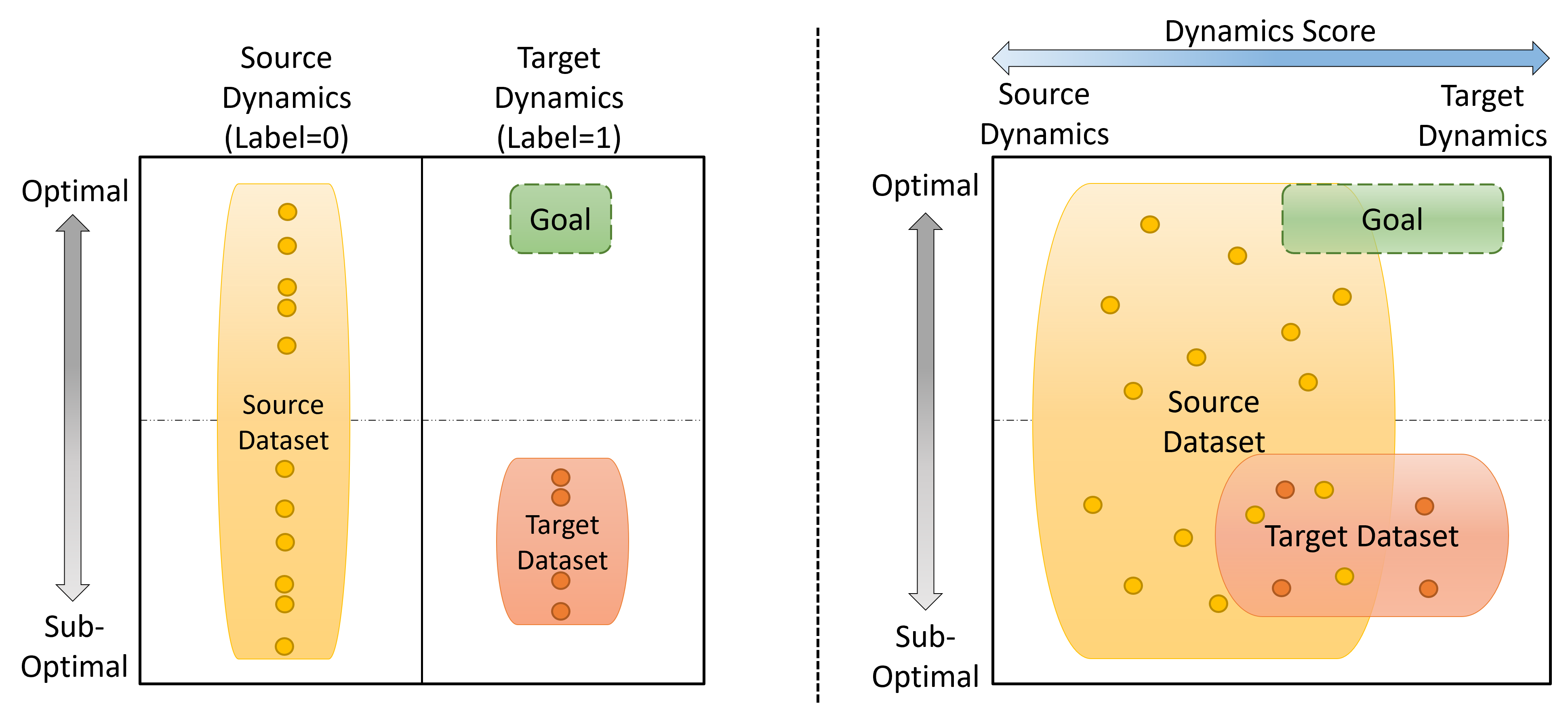}
    \caption{(Left) We utilize an accessible off-dynamics source dataset to enhance a limited target dataset for Offline RL. Our goal is to generate optimal trajectories within the green region. (Right) By conditioning a diffusion planner with our proposed continuous dynamics score, we enable the model to capture the underlying dynamics structure within the latent space through overlapping dynamics information.
    }
    \label{fig:high}
    \vspace{-10pt}
\end{figure}

Several approaches have been proposed to leverage off-dynamics samples to boost RL training. One popular method is to introduce a reward bonus to incentivize agents for taking actions which end up resembling the target environment \cite{eysenbach2020off,eysenbach2022mismatched, niu2022trust, liu2021dara, liu2021unsupervised}. However, this incurs extra cost and renders the learning of the primary task to be less effectively due to the embedded nature of the reward bonus. An alternative method is to apply transformations to the source domain \cite{hanna2017grounded,karnan2020reinforced,Christiano2016TransferFS}, making it behave like the target domain. This allows the agent to learn as if it were in the target domain. However, this  requires \textit{online interactions} with the target environment, making it unsuitable for our problem setting.

To address these limitations, we propose to utilize the flexibility and expressiveness of \textit{diffusion probabilistic models} (DPMs)  \cite{sohl2015deep, song2021scorebased, ho2020denoising} to learn a joint distribution of both source and target data for greater data efficiency. DPMs have demonstrated impressive capabilities in image generation \cite{saharia2022photorealistic, nichol2021glide}, audio generation \cite{kong2020diffwave}, video generation \cite{ho2022video} and more recently in Offline RL \cite{janner2022diffuser, ajay2023is, hu2023instructed}. 
To enable our model to generate trajectories for the target environment, we utilize classifier-free guidance \cite{Ho2022ClassifierFreeDG} by conditioning our model with dynamics-related contexts.
For this context, we propose a \textit{continuous dynamics score} as an alternative to the discrete dynamics labels as seen in Figure \ref{fig:high}. 
This ``soft score'' allows for greater coverage, enabling overlap between trajectories from different datasets.
Furthermore, we incorporate an inverse-dynamics context that measures the closeness to the target dynamics, ensuring the generated trajectories adhere to the target environment's dynamic constraints.
Together, the context facilitates capturing the underlying dynamics structure within the latent space more effectively, leading to improved generation performance.

We conduct comprehensive experiments to evaluate the effectiveness of our method on a diverse set of challenging off-dynamics settings. Empirical results demonstrate that our method outperforms several strong baselines. The simplicity of the proposed dynamics-related contexts coupled with the powerful capabilities of DPMs allow us to effectively leverage an accessible off-dynamics source dataset. Lastly, we demonstrate that by modifying the context, we can interpolate between source and target dynamics, making the model more robust to subtle shifts.

\section{Related Work}

\subsection{Off-dynamics Reinforcement Learning}
Coined by \cite{eysenbach2020off}, \textit{Off-dynamics RL} refers to a unique RL scenario where the transition dynamics in the source domain, used to train the policy, differ from those in the target domain. It can be applied to two settings, as described below. 

\noindent\textbf{Off-dynamics Online RL.} 
In the online setting, it is assumed that online samples from both domains are accessible, typically with a limited amount available in the target domain.
In \cite{eysenbach2020off}, a reward adjustment is proposed to encourage agents to favour transitions that are similar to the target dynamics when selecting actions during source domain training.
The reward adjustment is based on auxiliary classifiers that are trained separately to differentiate between source-domain and target-domain transitions.
This concept of utilizing dynamics classifiers has also been adapted to other off-dynamics problems such as model-based RL \cite{eysenbach2022mismatched}, inverse RL \cite{kang2021off} and unsupervised RL \cite{liu2021unsupervised}. 

A special case of off-dynamics RL involves the use of simulators as the source domain \cite{hanna2017grounded,karnan2020reinforced,Christiano2016TransferFS, desai2020imitation}. 
Instead of reward adjustments, these approaches \textit{ground} the simulator by applying a transformation on the source domain \cite{hanna2017grounded,karnan2020reinforced,Christiano2016TransferFS} to make it behave like the target domain. In this case, the agent can learn as if it were in the target domain.
These transformation models employ forward and/or inverse dynamics models of the source and/or target domain. 
Instead of learning any dynamics models, some approaches \cite{desai2020imitation, jiang2021simgan} involve using a classifier by redefining the task as an adversarial learning problem \cite{ho2016generative, goodfellow2014generative}.

\noindent\textbf{Off-dynamics Offline RL.} 
In the offline setting, 
Niu et al. \cite{niu2022trust} introduce a method that combines offline samples from the target domain with off-dynamics online samples from the simulator. Lastly, the most relevant work to our setting, \cite{liu2021dara}, exclusively uses offline samples from both the source and target domains. Both of these approaches closely align with the method described in \cite{eysenbach2020off}, applying a reward adjustment to the offline samples to facilitate learning.

\subsection{Diffusion Models}
\noindent\textbf{Diffusion Probabilistic Models}. These generative models \cite{sohl2015deep, song2021scorebased, ho2020denoising} have gained significant attention for their ability to produce high-quality, diverse samples across various domains such as images \cite{saharia2022photorealistic, dhariwal2021diffusion, nichol2021glide}, audio \cite{kong2020diffwave, popov2021grad} and video \cite{ho2022video}. They learn to gradually construct data by reversing a process that adds noise to the data over several steps. Some of the key benefits include: (1) enabling the creation of flexible models that can accommodate arbitrary data structures, (2) enabling training in a tractable manner, and (3) exhibiting high stability during training and lower susceptibility to mode collapse, unlike in Generative Adversarial Networks (GANs) \cite{arjovsky2017wasserstein, gulrajani2017improved}.

\noindent\textbf{Conditional DPMs.} 
To enhance the capabilities of DPMs, conditional DPMs are introduced through classifier-guided sampling \cite{dhariwal2021diffusion} and a classifier-free approach \cite{Ho2022ClassifierFreeDG}. They have been applied to a broad spectrum of tasks. A notable application involves generating images from textual descriptions \cite{nichol2021glide, saharia2022photorealistic, dhariwal2021diffusion}, where the generative model produces visuals based on the provided text input. Besides utilizing textual contexts for conditioning, other contexts have been explored for conditional image generation, e.g., low resolution contexts for super-resolution \cite{ho2022cascaded}, binary masks contexts for in-painting \cite{lugmayr2022repaint} and semantic masks, human poses contexts for ControlNet \cite{zhang2023adding}. Beyond image generation, conditional DPMs have also achieved success in various fields. e.g. 2D molecular graph contexts for 3D graph generation \cite{xu2021geodiff}, partial point-cloud observations contexts for point-cloud synthesis \cite{lyu2021conditional}, and linguistic features contexts for audio generation \cite{kong2020diffwave}.

\subsection{Diffusion-based Planners.}
Recently, diffusion-based planners \cite{janner2022diffuser, ajay2023is} utilize DPMs to generate trajectories which address the \textit{Offline RL} challenges as discussed in \cite{levine2020offline, chen2021decision}. 
By leveraging the powerful generative capabilities of DPMs, diffusion-based planners are able to outperform existing offline RL methods while having the additional benefits of specifying flexible constraints or composing multiple skills \cite{janner2022diffuser, ajay2023is, hu2023instructed}.
They have been adopted for various problem settings or tasks, such Multi-agent RL \cite{zhu2023madiff}, Meta-RL \cite{ni2023metadiffuser}, Multi-task RL \cite{he2023diffusion} and Safe RL \cite{zheng2024safe}.
To maximize the return of the generated trajectories in these diffusion-based planners, Diffuser \cite{janner2022diffuser} uses classifier-guided sampling \cite{dhariwal2021diffusion} whereas Decision-Diffuser \cite{ajay2023is} adopts a classifier-free approach \cite{Ho2022ClassifierFreeDG}. 

One limitation of diffusion-based planners is the long inference time, attributed to the slow sampling process of DPMs. Researchers have proposed several methods to accelerate this process \cite{dong2024diffuserlite, LIS317}.
While numerous follow-up works on diffusion-based planners exist, none specifically focus on utilizing an off-dynamics dataset to train the diffusion-based planners. 
The most relevant work to ours is \cite{ni2023metadiffuser}, where a diffusion-planner uses trajectories generated from a large number of off-dynamics environments for meta-training, aiming to generalize to the target domain. 
In contrast, \textit{our approach concentrates on adaptation rather than generalization, utilizing only a single source domain}.

\section{Preliminaries}

\subsection{Reinforcement Learning}
\noindent\textbf{Markov Decision Process.}
In RL \cite{sutton2018reinforcement}, an environment is characterized by a Markov Decision Process (MDP) defined as \( \mathcal{M} = (S, A, P, R, \gamma, d_0) \), where \( S \) and \( A \) represent the state and action spaces, respectively . The transition dynamics are denoted by \( P: S \times A \times S \rightarrow [0,1] \), and the reward function is represented by \( R: S \times A \times S \rightarrow \mathbb{R} \). The discount factor is a scalar \( \gamma \in [0, 1)\) representing how much the agent prioritizes immediate rewards over future rewards and \( d_0 \) defines the initial state distribution. An agent's policy \( \pi: S \rightarrow A \) induces a probability distribution over trajectories \( \tau = (s_t, a_t, r_t)_{t\geq0} \), where $s_t, a_t, r_t$ are the state, action and reward at timestep $t$, respectively.
For a planning horizon of $H$, the expected return for a trajectory is given by \( R(\tau) = \sum_{t=0}^{H} \gamma^t r_t \). The goal of RL is to find the optimal policy \( \pi^* \) that maximizes the expected cumulative discounted reward with the objective: $\pi^* = \arg\max_{\pi} \mathbb{E}_{\tau \sim p_{\pi}}[R(\tau)].$

\noindent\textbf{Offline RL.}
In contrast to traditional RL, the offline RL framework \cite{fu2020d4rl, levine2020offline} employs a static dataset \( \mathcal{D} = \{(s, a, r, s')\} \) for learning. Here, \(s'\) denotes the subsequent state following the application of action \(a\) from state \(s\), bypassing the need for data collection through direct environment interaction using a policy, \( \pi \).
This dataset is collected by an unknown behavior policy \( \mu \), and is utilized for the learning of a new policy entirely from \( \mathcal{D} \), without any online interactions with the environment. This resembles supervised learning, where \( \mathcal{D} \) serves as the training set. 
 A significant problem in offline RL is the distribution shift between the behavior policy and the learnt policy \cite{levine2020offline}, resulting in overly optimistic value estimates when evaluating unseen states or actions.
 To address this, prior works explicitly constrain the learnt policy to be close to the behavior policy \cite{fujimoto2019off,  kumar2019stabilizing}, or reduce the optimism of the value functions \cite{kumar2020conservative, kostrikov2021offline}. An alternative approach involves bypassing the value estimate by framing offline RL as a sequence modeling problem \cite{janner2021offline, chen2021decision, janner2022diffuser}.

\subsection{Diffusion Models}
\label{sec:Diffusion}

\noindent\textbf{Diffusion Probabilistic Models.}
DPMs \cite{sohl2015deep, song2021scorebased, ho2020denoising} form a class of generative models that learn data distributions \( q(x) \) through a process that inversely models the addition of noise. Specifically, data points sampled from \( p_{\text{data}}(x) \) undergo a forward noising sequence to produce a Markov chain \( x_0:K \) defined as \( x_k \sim \mathcal{N}(\sqrt{\alpha_k} x_{k-1}, (1 - \alpha_k)I) \), with \( \alpha_{0:K} \) being the noise schedule and $K$ being the number of diffusion steps. This chain is reversed by a variational process \( p_{\theta}(x_{k-1} | x_k) = \mathcal{N}(x_{k-1} | \mu_{\theta}(x_k, k), (1 - \alpha_k)I) \), beginning with \( x_K \sim \mathcal{N}(0, I) \) and iteratively denoised until \( x_0 \) is recovered. The procedure can be optimized through a surrogate loss \cite{ho2020denoising}: 
\begin{equation}
\label{loss:ddpm}
    \mathcal{L}(\theta) = \mathbb{E}_{k \sim [1,K], x_0 \sim q, \epsilon \sim \mathcal{N}(0,I)} \left\| \epsilon - \epsilon_{\theta}(x_k, k) \right\|^2 
\end{equation}
where the mean of the reverse Gaussian is given by \( \mu_{\theta}(x_k, k) = \frac{x_k - \sqrt{1 - \alpha_k} \epsilon_{\theta}(x_k, k)}{\sqrt{\alpha_k}} \), with \( \bar{\alpha}_k = \prod_{s=1}^k \alpha_s \). 

\noindent\textbf{Conditional Diffusion Models.}
The DPM method can be extended to a conditional generative model \( p_{\theta}(x_{t-1} | x_t, c) \) using \( c \) as an input context through classifier-free guidance \cite{Ho2022ClassifierFreeDG}.
During sampling, the predicted noise is adapted to a weighted combination of conditional and non-conditional sampling, $\hat{\epsilon}_\theta\left(x_t \mid c\right)=(1+w) \epsilon_\theta\left(x_t \mid c\right)+\epsilon_\theta\left(x_t \mid \emptyset\right)$, where \( \emptyset\) represents the null context. and \(w\) regulates the trade-off between sample quality and diversity by balancing the conditioned and unconditioned models. In practice, the unconditioned model is obtained by applying dropout on the context embedding. While training a conditional model with classifier-guided sampling \cite{dhariwal2021diffusion} is possible, classifier-free guidance provides advantages such as improved control over generation and superior performance \cite{nichol2021glide, ajay2023is}.

\noindent\textbf{Diffusion Planning}
By framing Offline RL as a sequence modeling problem, we can simplify this task as a supervised learning and planning framework \cite{janner2021offline}. Building on this concept, \textit{Diffuser} \cite{janner2022diffuser} is introduced as a diffusion-based trajectory planning model that utilizes expressive DPMs to model trajectories in the following form:
\begin{equation}
    \label{eqn:traj}
    \boldsymbol{\tau}=\left[\begin{array}{llll}
    \boldsymbol{s}_0 & \boldsymbol{s}_1 & \ldots & \boldsymbol{s}_H \\
    \boldsymbol{a}_0 & \boldsymbol{a}_1 & \ldots & \boldsymbol{a}_H
    \end{array}\right],
\end{equation}
where $H$ is the planning horizon. The model is optimized based on Equation \ref{loss:ddpm}, with $\epsilon_{\theta}(\tau_k, k)$ being modeled by U-Nets \cite{ronneberger2015u}, chosen for their non-autoregressive, temporally local, and equivariant characteristics. Conditional sampling is employed to generate trajectories that maximize return. By defining $\mathcal{O}_t$ as a binary random variable indicating the optimality of timestep \(t\) in a trajectory, where $p\left(\mathcal{O}_t=1\right)=\exp \left(\gamma^t r\left(\boldsymbol{s}_t, \boldsymbol{a}_t\right)\right)$, it is sufficient to guide the trajectory with the gradient of the return: $\nabla \mathcal{J} = \sum_{t=0}^T \nabla_{\mathbf{s}_t, \mathbf{a}_t} r\left(\mathbf{s}_t, \mathbf{a}_t\right)= \nabla_{\boldsymbol{\tau}} \log p\left(\mathcal{O}_{1: T} \mid \boldsymbol{\tau}\right).$
A separate model $\mathcal{J}_\phi$ is trained to predict the cumulative rewards, and the gradients of $\mathcal{J}_\phi$ are used to guide the trajectory following the classifier-guided sampling procedure \cite{dhariwal2021diffusion}. 
Following \textit{Diffuser}, \textit{Decision-Diffuser} \cite{ajay2023is} adopts a classifier-free approach \cite{Ho2022ClassifierFreeDG}, utilizing reward information as context. 
In this work, we adopt the classifier-free approach due to the ease of incorporating additional contexts beyond return.

\subsection{Problem Formulation: Off-Dynamics Offline RL}
In this paper, we aim to enhance a limited target dataset offline using an accessible off-dynamics source dataset.  We go beyond the standard offline RL framework of using a single fixed static offline dataset $\mathcal{D}_\text{target}$. Formally, similarly defined in \cite{liu2021dara}, we incorporate an additional source dataset $\mathcal{D}_\text{source} = {(s, a, r, s')}$, collected by another unknown behavior policy $\mu_\text{source}$, where $\mu_\text{source} \neq \mu_\text{target}$. A key distinction is that the source dataset is derived from an easily accessible source MDP $\mathcal{M}_\text{source}$, which exhibits different transition dynamics from the target MDP $\mathcal{M}_\text{target}$, i.e., $ \exists (s,a,s') : P_\text{source}(s'|s,a) \neq P_\text{target}(s'|s,a).$
We assume that $\mu_\text{source}$ is nearly optimal, and $\mathcal{D}_\text{source}$ sampled under $\mathcal{M}_\text{source}$ is abundant but off-dynamics. Conversely, the target dataset is presumed to have limited and suboptimal trajectories sampled under $\mathcal{M}_\text{target}$. Our goal is to enable the transfer of knowledge between the offline datasets $\mathcal{D}_\text{source}$ and $\mathcal{D}_\text{target}$ to diminish the data dependency of $\mathcal{D}_\text{target}$ for learning an optimal policy for $\mathcal{M}_\text{target}$.

\section{Approach}

We present our novel approach to improve a limited, sub-optimal target dataset, \( \mathcal{D}_\text{target} \), by leveraging a larger, diverse but off-dynamics source dataset, \( \mathcal{D}_\text{source} \). 
We achieve this by training a conditional DPM to learn the joint distribution of both datasets.
Naively training both datasets would be ineffective, as it would bias the model towards the source dynamics due to the larger size and optimality of \( \mathcal{D}_\text{source} \). 
This bias would cause the generated trajectories not to align with the target environment's dynamics.

To address this challenge, the model is conditioned on the following two dynamics-related contexts:
\begin{itemize}
    \item \underline{Continuous dynamics score}: This replaces discrete labels (Figure \ref{fig:high}) with a ``soft'' score, allowing for smoother transitions and overlap between trajectories from different datasets.
    \item \underline{Inverse-dynamics context}: This measures how closely generated trajectories align with the target environment's dynamics, ensuring the generated samples adhere to its specific constraints.
\end{itemize}
Although both contexts are related to dynamics information, they serve different purposes. The continuous dynamics score focuses on aligning the source and target dynamics, whereas the inverse dynamics context is used to generate feasible actions based on the target dynamics.
Together, these contexts enable the model to learn the joint distribution while adhering the inherent dynamics within the latent space.

\subsection{Dynamics Score Context} 
\label{sec:Dynamics}

We aim to introduce a context that can effectively capture both the differences and similarities between \( \mathcal{D}_\text{source} \) and \( \mathcal{D}_\text{target} \) in terms of dynamics information. 
A straightforward method is assigning the context as a discrete one-hot label. 
However, this does not provide sufficient information due to the potential overlaps in dynamics between the datasets. 
For instance, there could exist transitions in both \( \mathcal{D}_\text{source} \) and \( \mathcal{D}_\text{target} \) that are identical, indicating a shared dynamic characteristic across both domains. 
Given this, a discrete labeling system could fail to capture the nuanced differences and shared attributes adequately.

To address this limitation, we propose replacing the discrete context with a continuous score. This score should also be symmetric, ensuring equal representation for trajectories originating from \( \mathcal{D}_\text{source} \) and \( \mathcal{D}_\text{target} \).
To achieve this, for each trajectory $\tau$ in the form of Equation \ref{eqn:traj}, we define the dynamics score as
\begin{equation}
\begin{split}
    \label{eqn:dyn_score}
    c_\text{dyn\_score}(\tau) &= \frac{1}{\kappa H} \sum_{t=0}^{H-1} \left\{\log[P_\text{target}(s_t,a_t,s_{t+1}) + \epsilon] \right. \\
    &\qquad\quad\quad  - \left. \log[P_\text{source}(s_t,a_t,s_{t+1}) + \epsilon] \right\}.
\end{split}
\end{equation}
where \( P_\text{source} \) and \( P_\text{target} \) represent the probability of a given transition originating from the source and target datasets respectively, $\epsilon$ is a small value to prevent infinities and $\kappa$ is a scaling parameter to keep the score within $[-1, 1]$. 
We apply logarithmic scaling to enhance the dynamic range, particularly at the extreme values of the output probabilities.
Overall, a smaller score indicates trajectories follow source-like dynamics while a larger score implies target-like dynamics, and a score of zero suggests equal likelihood of both.
Details of modelling \( P_\text{target} \) will be elaborated in  Section \ref{sec:Practical}.

\subsection{Inverse-dynamics Context}
\label{sec:Inverse}
When attempting to maximize the out-of-distribution return context through the diffusion sampling process, the generated trajectory may not fully adhere to the dynamics constraints of the underlying MDP. To better enforce these constraints, one approach is to execute an inverse action based on the inverse dynamics to maintain the agent's dynamic constraints, as seen in \cite{liang2023adaptdiffuser}. Specifically, a separately trained inverse-dynamics model is used to predict the action from two consecutive states that best conform to the underlying MDP. Similar to this approach, we integrate the inverse-dynamics information into our method. The key difference, however, is that instead of using the inverse action as a post-processing method, we directly use it as a context for the conditional model. The rationale is as follows: if the inverse action is computed after the trajectory generation, the consecutive states could already violate the dynamics constraints, and the inverse model cannot recover the correct action. On the other hand, incorporating the inverse dynamics constraints during training compels the trajectory to adhere as closely as possible during the conditional sampling process, thus avoiding violation of the dynamics constraints in the first place. Formally, for each trajectory $\tau$ in the form of Equation \ref{eqn:traj}, the inverse context, $c_\text{inverse}$ is computed as follows:

\begin{equation}
    \label{eqn:inverse}
    c_\text{inverse}(\tau) = \frac{1}{H} \sum_{t=0}^{H-1} \text{log} [ 1+ \|I_\text{target}(s_t, s_{t+1})-a_t\|_2 ],
\end{equation}
where $I_\text{target}$ is the target inverse dynamics model, $H$ is the planning horizon.
The logarithm function helps to compress the range of the large-valued errors.

\subsection{Practical Algorithm}
\label{sec:Practical}
We begin by outlining the process of learning \( P_\text{target} \), \( P_\text{source} \) and \( I_\text{inverse} \).
To model \( P_\text{target} \), we parameterize a binary classifier, \( p_\phi = p(s_t, a_t, s_{t+1}; \phi) \), with a multi-layer perceptron (MLP). For \( P_\text{source} \), we simply use \( 1-p_\phi \). We fit $p_\phi$ by minimizing the standard binary cross-entropy loss:
\begin{equation}
\begin{split}
    \label{loss:classify}
    & L(\phi) =
     - \mathbb{E}_{(s_t,a_t,s_{t+1})\sim \mathcal{D}_\text{source} \cup \mathcal{D}_\text{target}} \\
     & \qquad\qquad\qquad \left[ y \log(p_\phi) + (1 - y) \log(1 - p_\phi) \right], 
\end{split}
\end{equation}
where \( y=0,1 \) represents the source and target labels respectively. 

Next, to model the inverse dynamics \( I_\text{target} \), we parameterize it by $\psi$ with another MLP, and minimize the standard mean-squared error between the predicted action and true action $a_t$:
\begin{equation}
    \label{loss:inverse}
    L(\psi) = \mathbb{E}_{(s_t,a_t,s_{t+1})\sim \mathcal{D}_\text{target}} \left[ (I_\text{target}(s_t, s_{t+1}; \psi) - a_t)^2 \right].
\end{equation}

\begin{algorithm}[t]
\small
\caption{Training: Off-dynamics Conditional Diffusion Planners}
    \begin{algorithmic}[1]
        \State \textbf{Input:} Target dataset $\mathcal{D}_\text{target}$, Source dataset $\mathcal{D}_\text{source}$
        \State \textbf{Input:} Number of training updates $N$, \\ \;\;\;\;\;\;\;\;\;\;\;Number of diffusion time steps $T$
        \State Initialize dynamics score model $q_\phi$ and inverse model $f_\psi$
        \State Fit $q_\phi$ using Loss in Eq.\ref{loss:classify} over $\mathcal{D}_\text{target}\cup \mathcal{D}_\text{source}$
        \State Fit $f_\psi$ using Loss in Eq.\ref{loss:inverse} over $\mathcal{D}_\text{target}$
        
        \State \text{// Start Conditional Diffusion Training}
        \State Initialize Conditional U-Nets $\epsilon_\theta$ 

        \For{$n$ in $1, 2, ..., N$}
            \State Sample stratified batch $\tau_\mathcal{B} \in \mathcal{D}_\text{target} \cup \mathcal{D}_\text{source}$
            \State Compute context $c_{dyn\_score}(\tau_\mathcal{B}) = q_\phi(\tau_\mathcal{B})$
            \State Compute context $c_{inverse}(\tau_\mathcal{B}) = f_\psi(\tau_\mathcal{B})$
            \State Set full context $\boldsymbol{y}(\tau_\mathcal{B}) = [ R(\tau_\mathcal{B}), c_{dyn\_score}(\tau_\mathcal{B}), c_\text{inverse}(\tau_\mathcal{B})].$
            \For{$t$ in $1, 2, ..., T$}
                \State Update $\theta$ with $\epsilon_\theta(\tau_\mathcal{B}, t, \boldsymbol{y}(\tau_\mathcal{B}))$  using Loss in Eq.\ref{loss:ddpm_cond}
            \EndFor
        \EndFor

    \end{algorithmic}
    \label{algo:1}
\end{algorithm}

Now, we introduce a practical algorithm that combines all components for training a off-dynamics conditional diffusion-based planner. 
Our model follows the classifier-free approach with input trajectories $\tau$, following Equation \ref{eqn:traj}.
The models are conditioned on the full context $\boldsymbol{y}(\tau)$, which consists of the dynamics score, the inverse dynamics and the normalised return $R(\tau) \in [0,1]$ as follows: 
\begin{equation}
    \label{eqn:context}
    \boldsymbol{y}(\tau) = [ R(\tau), c_{dyn\_score}(\tau), c_\text{inverse}(\tau)].
\end{equation}
This context summarizes the optimality and the dynamics information of each trajectory in a continuous form as movitated in Fig. \ref{fig:high}. 
By leveraging dynamics information, the conditional model gains the ability to flexibly learn from both \( \mathcal{D}_\text{source} \) and \( \mathcal{D}_\text{target} \) with greater data efficiency. It also enables the conditional model to generate optimal trajectories from either domain in a seamless manner. The objective for the conditional diffusion process is,
$$\max _\theta \mathbb{E}_{\tau \sim (\mathcal{D}_\text{source}\cup \mathcal{D}_\text{target})}\left[\log p_\theta\left(\boldsymbol{x}_0(\tau) \mid \boldsymbol{y}(\tau)\right)\right]$$
with the loss given by
\begin{equation}
\label{loss:ddpm_cond}
    \mathcal{L}(\theta) = \mathbb{E}_{k \sim [1,K], x_0 \sim q, \epsilon \sim \mathcal{N}(0,I)} \left\| \epsilon - \epsilon_{\theta}(x_k(\tau), k, y(\tau)) \right\|^2 
\end{equation}
The complete training procedure is detailed in Algorithm \ref{algo:1}. During planning, we set the target context $y(\tau) = [1, 1, 0]$ to generate trajectories that maximize the reward for the target environment. This aligns with Equations \ref{eqn:dyn_score} and \ref{eqn:inverse}, where $P_\text{target}(s_t, a_t, s_{t+1}) = 1$ and $I_\text{target}(s_t, s_{t+1}) = a_t$, ensuring the generated trajectories adhere to the target dynamics.

\begin{table*}[t]
\centering

\begin{adjustbox}{width=\textwidth,center}
\begin{tabular}{cccc|c|c|ccc|ccc}
\Xhline{1pt}  
\multirow{2}{*}{Environment} & \multirow{2}{*}{Property} & \multirow{2}{*}{Source} & \multirow{2}{*}{Target} & $\mathcal{D}_\text{source}$ & $\mathcal{D}_\text{target}$ & \multicolumn{3}{c|}{$\mathcal{D}_\text{source}$ pretrain + $\mathcal{D}_\text{target}$ finetune} & \multicolumn{3}{c}{$\mathcal{D}_\text{source}\cup\mathcal{D}_\text{target}$ Joint Training} \\ \cline{5-6} \cline{6-7} \cline{8-9}  \cline{9-12} 
 &  & &  & Diffuser & Diffuser & Diffuser & CQL & BCQ & CQL+DARA & BCQ+DARA & Proposed \\ \hline
\multirow{3}{*}{Half-cheetah} & total mass & 14 & 7 & 30.2 & 30.5 & 32.6 & 26.3 & 28.7 & 27.3 & 19.8 & \textbf{56.8} $\pm$ 5.9 \\
 & torso size & 0.046 & 0.092 & 43.6 & 43.9 & 38.2 & 32.0 & 46.8 & 45.9 & 32.2 & \textbf{54.1} $\pm$ 7.9 \\
 & control range & [-1,1] & [-0.5,0.5] & 41.7 & 43.1 & 25.2 & 52.2 & 2.2 & 53.2 & 53.1 & \textbf{61.5} $\pm$ 6.6 \\ \hline
\multirow{3}{*}{Walker2d} & thigh action& Enabled & Disabled & 22.6 & 24.4 & 39.1 & 28.8 & -0.3 & 32.2 & 10.2 & \textbf{58.5} $\pm$ 9.0 \\
 & foot gear torque & 100 & 70 & 42.9  & 43.1 & 36.7 & 46.8 & 51.0 & -0.2 & 44.2 & \textbf{63.4} $\pm$ 12.2 \\
 & foot length & 0.1 & 0.25 & 42.7 & 42.9 & 41.6 & 48.3 & 43.6 & -0.2 & 5.3 & \textbf{63.0} $\pm$ 10.7 \\ \hline
\multirow{3}{*}{Hopper} & foot friction & 2.0 & 1.7 & 50.7 & 51.3 & 52.5 & 32.1 & 40.2 & 49.7 & 109.5 & \textbf{124.2} $\pm$ 31.5 \\
 & torso stiffness & 0 & 3 & 73.9 & 75.8 & 72.2 & 65.7 & 72.5 & 51.8 & 42.9 & \textbf{95.0} $\pm$ 6.9 \\
 & leg size & 0.03 & 0.04 & 74.2 & 76.0 & 57.3 & 36.8 & 82.8 & 75.6 & 93.1 & \textbf{95.9} $\pm$ 9.3\\ \hline 
 & & & \textbf{Average}  & 46.9 & 47.9 & 43.9 & 41.0 & 40.8 & 37.3 & 45.6 & \textbf{74.7} \\ 
\cmidrule[1pt]{4-12} 
\end{tabular}
\end{adjustbox}
\caption{\textbf{Mean normalized scores evaluated on the target environment over 900 episodes across 9 diverse settings.} The columns `Source' and `Target' represent the dynamics settings of $\mathcal{D}_\text{source}$ and $\mathcal{D}_\text{target}$, respectively. Our proposed method is a conditional diffusion planner with contexts according to Equation \ref{eqn:context}, trained with Algorithm \ref{algo:1}.
}
\label{table1}
\end{table*}

\section{Experiments}

We design and conduct comprehensive experiments to thoroughly compare the effectiveness of our method with existing ones. We begin by outlining the experimental setup, which involves creating a diverse and challenging set of off-dynamics datasets for training and evaluation. Nine distinct settings are used to compare our approach with several strong baselines. Additionally, an ablation study is conducted to investigate the specific contribution of each dynamics context to the performance improvement. Finally, we investigate the robustness and capacity of our model to seamlessly interpolate between source and target dynamics.

\subsection{Experimental Setup}
\label{sec:experiment-setup}

\noindent\textbf{Datasets.}
We formulate our experimental datasets based on our proposed framework, illustrated in Figure \ref{fig:high} (left). 
We require a large and diverse offline source dataset, and select the Hopper, Walker2d, and Halfcheetah medium-expert datasets from D4RL \cite{fu2020d4rl}, each consisting of around 2 million samples. For the target dataset, we need limited and sub-optimal samples with different dynamics compared to the source. To achieve this, we first create multiple off-dynamics variants by modifying parameters like mass, size, control range, friction, and gear torques within each environment. Subsequently, we collect 10,000 samples from each modified environment using a behavioral policy trained with Diffuser in the respective source environment. This diverse and challenging off-dynamics setting allows for comprehensive evaluation of our method.

\noindent\textbf{Baselines.} We benchmark against various data-driven control algorithms for off-dynamics offline RL. We apply the off-dynamics reward compensation method DARA \cite{liu2021dara} to several well-performing offline RL algorithms. Among these offline RL algorithms, we include model-free methods like BCQ \cite{fujimoto2019off} and CQL \cite{kumar2020conservative}, as well as diffusion-based methods like Diffuser \cite{janner2022diffuser}. In addition to DARA, we include the fine-tuning method, which performs 10,000 additional updates on a pre-trained source model using the target dataset. 

\noindent\textbf{Implementation details.} For the hyper-parameters of DPM, we follow the default settings in Diffuser \cite{janner2022diffuser}. For both the classifier and inverse models, we use basic MLPs with 2 hidden layers and 32 nodes, each trained for 200k updates. We apply \textit{min-max normalization} to the outputs of these models prior to context computations. During training, we apply a context \textit{dropout} of $p=0.5$ for each context independently. To address the imbalanced dataset, we ensure a 1:1 ratio of source and target data in each batch during sampling. 
During sampling, we use a conditional sampling weight of $w=0.9$ for all environments. 
We evaluate our models over 300 episodes across 3 seeds (total 900 episodes) and compute the normalized score over the target environments \cite{fu2020d4rl}.

\subsection{Main Results}
We conduct experiments on three different environments, each with three variations in their physical properties, resulting in a total of nine different settings. Table \ref{table1} displays the normalized scores of different methods over four training types: 1) source only, 2) target only, 3) source pretraining with target finetuning, and 4) joint source and target training. For the source only setting, which uses a Diffuser trained on optimal samples from the source dataset (``$\mathcal{D}_\text{source}$'' column), the performance is significantly decreased across all environments when evaluated on the target environment. This finding highlights the inherent challenges presented by our off-dynamics environments. We use this behavioral policy to collect samples for our target dataset. 
Next, for the target-only setting, we trained a model with Diffuser using exclusively target samples (``$\mathcal{D}_\text{target}$'' column). This resulted in a slight improvement over the model trained solely on source data. This score serves as a baseline for the target dataset's performance.

Moving on to finetuning (``$\mathcal{D}_\text{source}$ pretrain+$\mathcal{D}_\text{target}$ finetune'' column), we observe that all three methods (Diffuser, CQL, BCQ) generally perform worse than the source-only Diffuser. A possible reason is that the model struggles to balance learning dynamics from the target dataset while retaining the knowledge of optimal behaviors learned from the source dataset during the finetuning process. Lastly, in the joint training setting (``$\mathcal{D}_\text{source}\cup\mathcal{D}_\text{target}$ Joint Training'' column), both DARA methods face difficulties in this low-data regime despite the dynamics-aware reward adjustment. On the other hand, our proposed conditional DPM with dynamics contexts outperform all baselines by a significant margin. This is attributed to the powerful DPM's ability to generate high-reward trajectories for the target environment, guided by the dynamics-related contexts. Below we break down the contribution of each individual context.

\begin{figure}
    \centering
    \includegraphics[width=\linewidth]{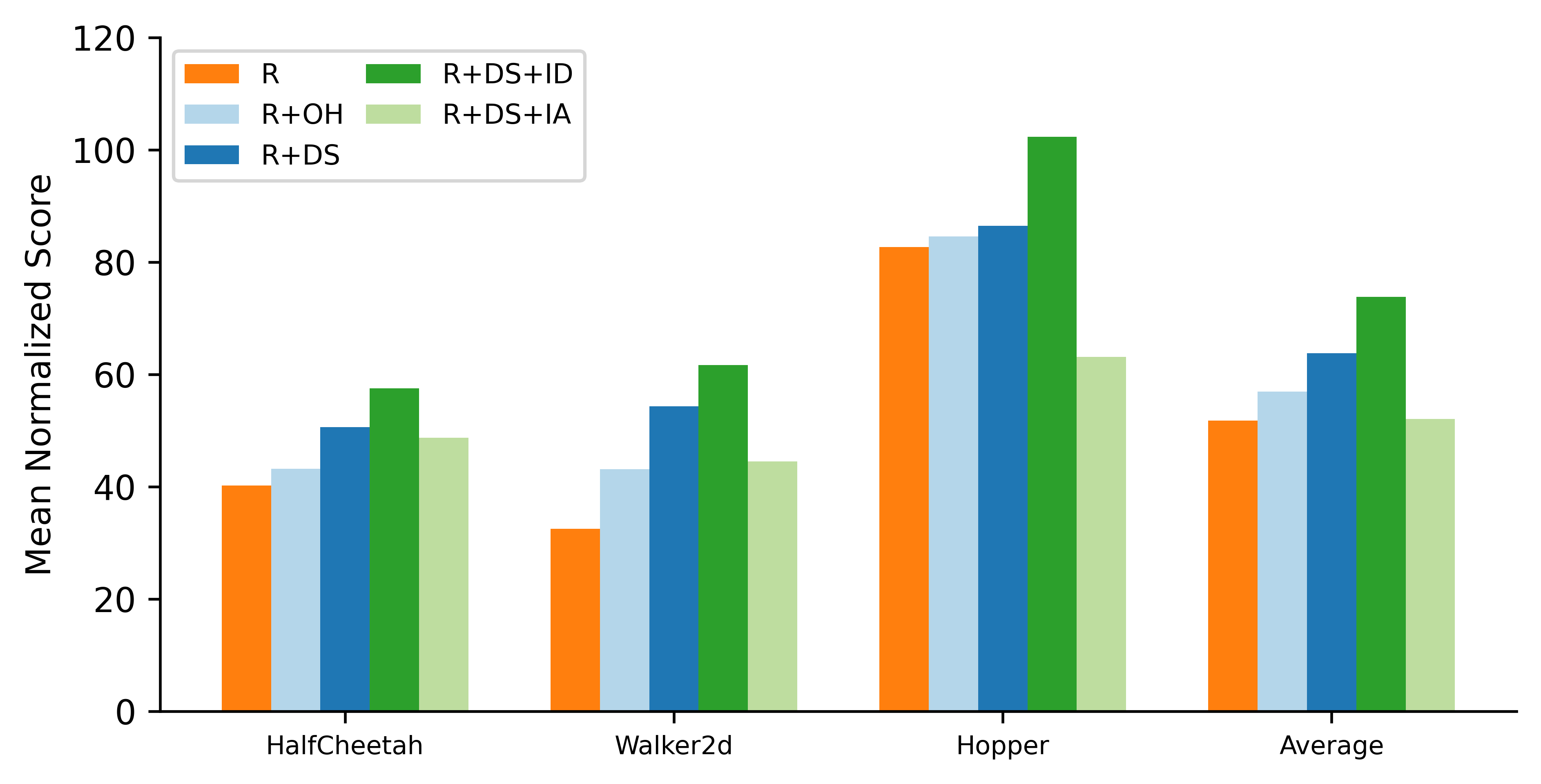}
    \caption{ \textbf{Ablation: models trained with different contexts following Algorithm \ref{algo:1}.} We report the mean normalized score across different settings per environment. (`R', Orange) represents the base model conditioned on only the return. (`R+OH', LightBlue) adds on the one-hot source/target label as contexts to base model. (`R+DS', Blue) adds on the dynamics score as contexts to base model. (`R+DS+ID', Green) further adds on the inverse-dynamics context. (`R+DS+IA', LightGreen) applies inverse action on `R+DS'. 
    }
    \label{fig:ablation}
    \vspace{-10pt}
\end{figure}

\subsection{Contexts Analysis}

To better understand each context's contribution, we conducted additional experiments and present the mean normalized scores per environment in Figure \ref{fig:ablation}.

\textbf{Source/Target Dynamics}
We evaluated the impact of using discrete labels (`R+OH', LightBlue) versus our proposed soft score (`R+DS', Blue), as detailed in Section \ref{sec:Dynamics}.  While discrete labels showed some improvement over the baseline ('R', Orange), our soft score consistently outperformed both, confirming its superior ability to capture the underlying dynamics structure.

\textbf{Inverse Dynamics}
Incorporating the inverse dynamics context (`R+DS+ID', Green) from Section \ref{sec:Inverse}, in addition to the dynamics score, yielded further improvements. This suggests that the inverse dynamics context provides valuable guidance during sampling, leading to even better adherence to target dynamics.
However, directly applying the inverse action on R+DS proved detrimental (`R+DS+IA', LightGreen), underlining our hypothesis that consecutive states generated by R+DS might violate dynamics constraints, hindering the inverse model's ability to recover correct actions. This was particularly pronounced with Hopper, where using the inverse action led to performance worse than relying solely on the return context. 

In summary, these ablation studies highlight the importance of both our soft score for dynamics alignment and the inverse dynamics context for generating feasible actions. They also demonstrate the limitations of directly applying inverse actions in the presence of potential constraint violations.

\begin{figure}[t]
    \centering
    \includegraphics[width=\linewidth]{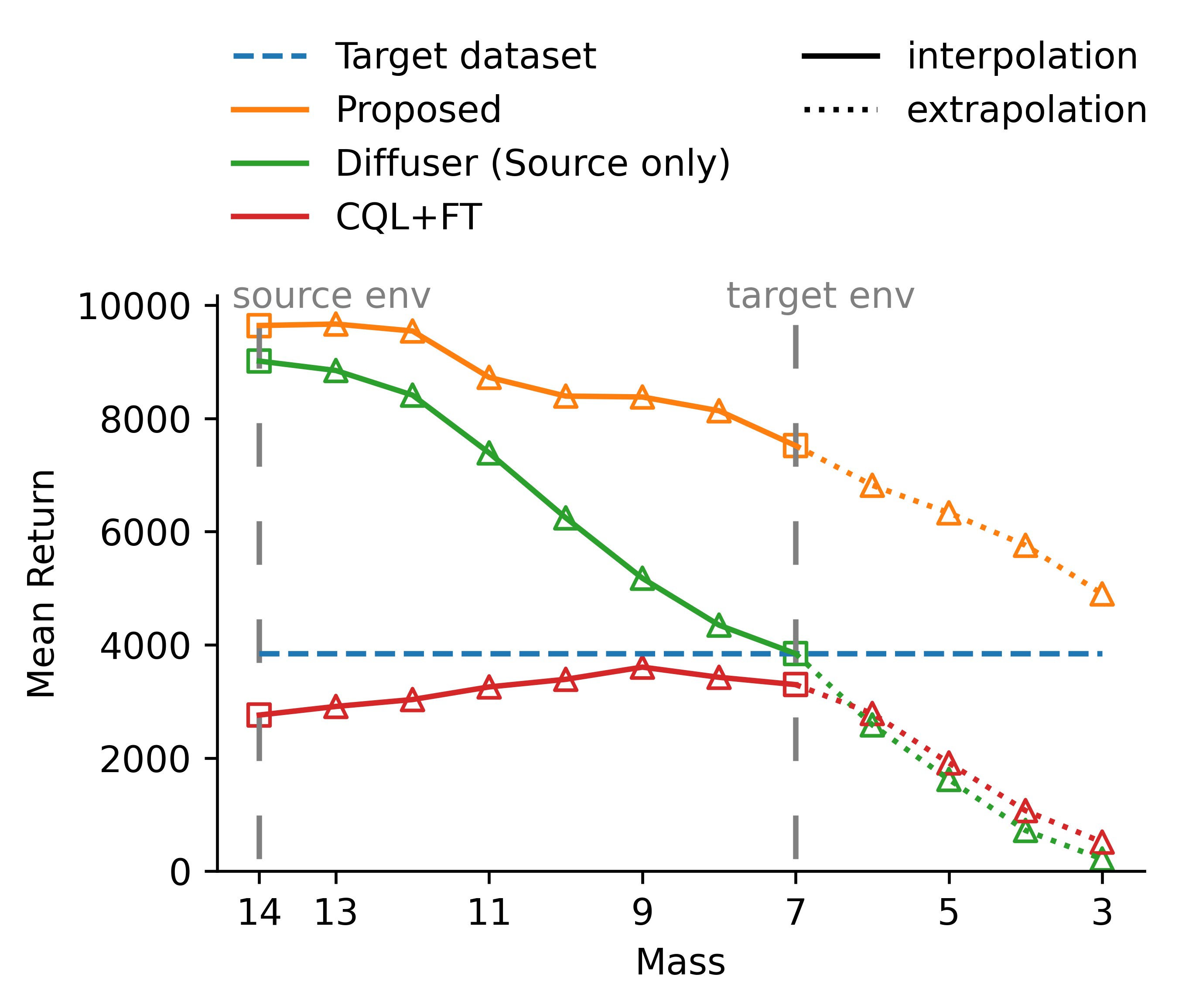}
    \caption{ \textbf{Plot of the generalisation capabilities for Halfcheetah.} Models are trained on $\mathcal{D}_\text{source}$ ($m=14$), $\mathcal{D}_\text{target}$ ($m=7$). Models are evaluated at interpolated masses $8\leq m \leq13$ and extrapolated masses $3 \leq m \leq 6$. Mean returns are shown due to varying normalizing score factors across different masses. 
    }
    \vspace{-10pt}
    \label{fig:interpolation}
\end{figure}

\subsection{Robustness}
Inspired by the success of diffusion models in other domains, where they achieve smooth interpolation within the latent space by controlling contexts, we investigate whether our model exhibits similar behaviors. This ability becomes crucial in real-world scenarios, as target dynamics often undergo subtle shifts.
To explore this, we leverage one of our experimental settings involving Halfcheetah with varying masses (Table \ref{table1}, first row). The source and target datasets correspond to masses of $m=14$ and $m=7$, respectively. 
Using the models trained with these two datasets, we further evaluate the masses in between (interpolation: $8\leq m \leq13$) and beyond (extrapolation: $3 \leq m \leq 6$). 
We evaluate our model, alongside two baselines: (1) Diffuser, using only source data; (2) CQL+FT, source pre-train with target fine-tuning.
For our approach, we apply a simple linear scale on the dynamics-related contexts based on mass for intermediate evaluations. During extrapolation $3 \leq m \leq 6$, we utilize the context corresponding to the target mass $m = 7$. 

Figure \ref{fig:interpolation} presents the results.
Our method (orange line) exhibits robust performance across the interpolated range ($8\leq m \leq13$), even without training data following these environment's dynamics. Compared to CQL+FT (red line), the model maintains similar performance for $m=8,9$, but exhibits a gradual decline beyond those values. This aligns with the expectation that the fine-tuned model has forgotten most of its source information. 
While sharing a similar architecture, the Diffuser trained on source data (green line) experiences a rapid performance drop beyond the source mass of $m = 7$. This suggests that the dataset with matching dynamics, despite its small quantity, remains crucial for effectively training our model.
Finally, when extrapolating masses outside the range of the training data ($3 \leq m \leq 6$), all three models exhibit similar limitations in their capabilities. This is evident in the similar rate of performance degradation observed for all models in this region.

\section{Conclusion}
In this paper, we propose a new approach, which utilizes a conditional DPM with dynamics-related contexts to address the challenge of data scarcity in offline RL. We introduce a continuous dynamics score and an inverse-dynamics context to effectively capture the underlying dynamics structure within the latent space, enabling the model to learn from both a larger off-dynamics source dataset and a limited, sub-optimal target dataset. Experimental results demonstrate that our method significantly outperforms various baselines. Ablation studies further reveal the critical role of each dynamics context in improving performance. Additionally, our model exhibits promising robustness in handling interpolation scenarios, showcasing its potential for real-world applications with dynamic shifts in the target environment. Our future work aims to expand the applicability of our method by incorporating multiple off-dynamics datasets, enabling adaptation to changes in robot embodiment, and ultimately validating its effectiveness in real-world robotics scenarios.

\bibliographystyle{IEEEtran}
\bibliography{IEEEexample}

\end{document}